\documentclass[conference, letterpaper]{IEEEtran}

\usepackage[T1]{fontenc}
\usepackage{aecompl} 
\usepackage{caption}

\ifCLASSINFOpdf
   \usepackage[pdftex]{graphicx}

   \graphicspath{{{\vartheta}_{k}^v/pdf/}{{\vartheta}_{k}^v/jpeg/}}

   \DeclareGraphicsExtensions{.pdf,.jpeg,.png}
\else

   \usepackage[dvips]{graphicx}

   \graphicspath{{{\vartheta
}_{k}^v/eps/}}
  
   \DeclareGraphicsExtensions{-eps-converted-to.pdf}
\fi

\usepackage{amsmath}
\usepackage{tabularray}
\usepackage{subeqnarray}
\usepackage{cases}
\usepackage{cite}
\usepackage{graphicx}
\usepackage{subfigure}
\usepackage{tabularx,booktabs}
\usepackage{dingbat}
\usepackage{diagbox}
\usepackage[ruled,linesnumbered]{algorithm2e}
\usepackage{amssymb}
\usepackage{algorithmic}
\usepackage{esint}
\usepackage{color}
\usepackage{lipsum}
\usepackage{cuted}
\usepackage{stfloats}
\usepackage{ntheorem}
\usepackage{threeparttable}
\usepackage{bbm}
\usepackage{tabu}
\usepackage{multirow}
\usepackage{makecell}
\usepackage{colortbl}
\usepackage{url}
\usepackage{array}
\usepackage[colorlinks,
            linkcolor=black,       
            anchorcolor=black,  
            citecolor=black,
            urlcolor=black,        
            ]{hyperref}
\usepackage{bm}
\newcolumntype{C}{>{\centering\arraybackslash}X}
\theoremseparator{:}

\usepackage[cmintegrals]{newtxmath}

\begin{document}

\title{Channel-Aware Throughput Maximization for Cooperative Data Fusion in CAV}
\vspace{-2.5cm}
\author{\IEEEauthorblockN{Haonan An\IEEEauthorrefmark{2},
Zhengru Fang\IEEEauthorrefmark{2}, Yuang Zhang\IEEEauthorrefmark{3}, Senkang Hu\IEEEauthorrefmark{2}, Xianhao Chen\IEEEauthorrefmark{4}, Guowen Xu\IEEEauthorrefmark{2} and Yuguang Fang\IEEEauthorrefmark{2}}\\
\IEEEauthorblockA{\IEEEauthorrefmark{2}Department of Computer Science, City University of Hong Kong, Hong Kong}\\
\vspace{-3mm}
\IEEEauthorblockA{\IEEEauthorrefmark{3}Department of Automation, Tsinghua University, China}\\
\vspace{-3mm}
\IEEEauthorblockA{\IEEEauthorrefmark{4}Department of Electrical and Electronic Engineering, the University of Hong Kong, Hong Kong}}



\maketitle
\newcommand\blfootnote[1]{%
\begingroup
\renewcommand\thefootnote{}\footnote{
#1}%
\addtocounter{footnote}{-1}%
\endgroup
}


\vspace{-0.4cm}
\begin{abstract}
Connected and autonomous vehicles (CAVs) have garnered significant attention due to their extended perception range and enhanced sensing coverage. To address challenges such as blind spots and obstructions, CAVs employ vehicle-to-vehicle (V2V) communications to aggregate sensory data from surrounding vehicles. However, cooperative perception is often constrained by the limitations of achievable network throughput and channel quality. In this paper, we propose a channel-aware throughput maximization approach to facilitate CAV data fusion, leveraging a self-supervised autoencoder for adaptive data compression. We formulate the problem as a mixed integer programming (MIP) model, which we decompose into two sub-problems to derive optimal data rate and compression ratio solutions under given link conditions. An autoencoder is then trained to minimize bitrate with the determined compression ratio, and a fine-tuning strategy is employed to further reduce spectrum resource consumption. Experimental evaluation on the OpenCOOD platform demonstrates the effectiveness of our proposed algorithm, showing more than 20.19\% improvement in network throughput and a 9.38\% increase in average precision (AP@IoU) compared to state-of-the-art methods, with an optimal latency of 19.99 ms.
\end{abstract}
\IEEEpeerreviewmaketitle
\section{Introduction}
\IEEEPARstart{R}{ecently}, autonomous driving has emerged as a promising technology for smart cities. By leveraging communication and artificial intelligence (AI) technologies, autonomous driving can significantly enhance the performance of a city's transportation system. This improvement is achieved through real-time perception of road conditions and precise object detection from onboard sensors (such as radars, LiDARs, and cameras), thereby improving road safety without human intervention \cite{chen2023vehicle}. Moreover, the ability of autonomous vehicles to adapt to dynamic environments and communicate with surrounding infrastructure and vehicles is crucial for maintaining the timeliness and accuracy of collected data, thereby enhancing the overall system performance.

Joint perception among connected and autonomous vehicles (CAVs) is a key enabler to overcome the limitations of individual agent sensing capabilities \cite{li2023learning}. Specifically, cooperative CAVs enable a CAV to have a longer perception range and avoid blind spots caused by occlusions. Compared to individual vehicle perception, the advantage of collaborative perception lies in enhancing observations from different perspectives and extending the perception range beyond line of sight, up to the maximum sensing boundary within all CAVs \cite{9387631}. There are three methods for information fusion through V2V communications: (1) early fusion for raw sensed data, (2) intermediate fusion for intermediate features from a deep learning model, and (3) late fusion for detection results \cite{9165167}. Recent state-of-the-art \cite{xu2022opv2v} indicates that intermediate fusion is the trade-off between perception accuracy and bandwidth requirements.

However, due to the large amount of sensed data (e.g., point clouds and image sequences), data transmissions for CAVs in cooperative perception require massive network throughput, whereby the limitation of capacity results in communication bottlenecks. According to the KITTI dataset \cite{6248074}, each frame generated by 3-D Velodyne laser scanners consists of 100,000 points, while the smallest recorded scene comprises 114 frames, amounting to over 10 million points. Therefore, it is impractical to transmit such massive amounts of data by V2V networks among large-scale vehicular nodes. To overcome the challenge of network throughput, existing studies have attempted to use either a communication-efficient collaborative perception framework \cite{Where2comm, hu2022where2comm}, a point cloud feature-based perception framework \cite{chen2019f, chen2020pointcloud}, or a frame for sending compressed deep feature maps \cite{wang2020v2vnet}.

Collaborative sensing capabilities are not limited to road vehicles but extend to various autonomous platforms, such as underwater robots and unmanned surface vessels, which also share sensory data to improve perception and decision-making accuracy . These platforms can take advantage of wireless communication technologies to improve cooperative navigation and perception, particularly in harsh environments. Cooperative object classification \cite{arnold2021data} is another important aspect of collaborative sensing that improves the identification of objects through data sharing among multiple autonomous agents.

In terms of wireless communication, the challenges in providing stable and reliable connections among moving vehicles are crucial, particularly under dynamic and uncertain channel conditions. Machine learning has played an important role in enabling adaptive wireless communication, where learning-based methods such as federated learning, split learning, and edge intelligence have been leveraged to improve the efficiency of communication among CAVs and other autonomous systems \cite{qu2024mobileedgeintelligencelarge, qu2024trimcachingparametersharingedgecaching}.

For collaborative perception in autonomous driving, there have been recent advancements focusing on improving coordination among connected vehicles to enhance perception accuracy, especially under real-world constraints like limited bandwidth and high mobility \cite{hu2024collaborativeperceptionconnectedautonomous, fang2024pacp, hu2024agentscodriverlargelanguagemodel, 10610199, tao2024directcpdirectedcollaborativeperception, ren2024collaborative, 9978606, han2023collaborative, xing2021toward, wang2023umc, xiao2023toward, wang2023core, xiang2023hm, su2023uncertainty}. Visual data collected by multiple cameras from different vehicles provide a richer set of observations, enabling more reliable detection of objects and hazards. However, it also introduces significant communication overhead due to the large size of image data, motivating the development of efficient collaborative perception frameworks for visual data \cite{fang2024prioritized, fang2024pib, li2023multi}.

Attention mechanisms, such as those used in Transformer models, have proven effective for capturing relationships in time-series data, making them particularly useful in collaborative perception for autonomous driving \cite{vaswani2017attention}. Dual-stage attention-based recurrent neural networks \cite{qin2017dual} and multi-time attention networks \cite{shukla2021multi} have demonstrated the ability to handle irregularly sampled data, which is often the case in vehicular sensor networks due to unpredictable communication delays and packet loss. Moreover, interpolation-prediction networks have been proposed to address challenges in irregularly sampled time series by learning both interpolation and prediction simultaneously \cite{shukla2019interpolation}, which is relevant for fusing data from multiple CAVs with varying data rates.

Dynamic generative models, such as dynamic Gaussian mixture-based deep generative models, have also been proposed to improve forecasting and compression in sparse multivariate time series, making them highly suitable for the dynamic environments faced by CAVs \cite{wu2021dynamic}. Additionally, set functions for time series \cite{horn2020set} provide a novel way to represent and process data collected from multiple vehicles, ensuring efficient handling of temporal dependencies. Graph-guided networks for irregularly sampled time series \cite{zhang2022graph} further enhance the capability to model complex relationships in vehicular networks, leading to more efficient collaborative perception.

SLAM (Simultaneous Localization and Mapping) is another critical component in autonomous driving systems, providing a foundation for localizing vehicles and mapping their surroundings. Surveys on SLAM highlight the importance of robust techniques for mapping and localization \cite{aulinas2008slam}. Probabilistic data association methods for SLAM are essential for achieving semantic localization in dynamic environments \cite{bowman2017probabilistic, atanasov2016localization}. Semantic mapping is crucial for data association in SLAM, ensuring robust data fusion from multiple sources \cite{bernreiter2019semantic}. Dynamic visual SLAM, when combined with deep learning, further improves the accuracy and robustness of SLAM in changing environments \cite{beghdadi2022overview}.

Object tracking is another important aspect of autonomous driving, enabling vehicles to maintain awareness of surrounding objects over time. Simple online and real-time tracking (SORT) has been proposed as an effective approach for achieving this \cite{bewley2016simple}. Simulators such as CARLA \cite{dosovitskiy2017carla} have played a significant role in testing and validating autonomous driving systems, providing a controlled environment to evaluate new algorithms under different traffic scenarios. Object detection methods, like PointPillars \cite{lang2019pointpillars}, are also crucial for processing point cloud data efficiently, which is fundamental for perception in autonomous driving.

For cooperative perception among autonomous agents, effective communication strategies are necessary to ensure spatial coordination and maximize perception accuracy. Multi-agent spatial coordination techniques \cite{glaser2021overcoming1, glaser2021overcoming2} are key in overcoming obstructions and improving information sharing among vehicles. Furthermore, feature-level consensus is critical for ensuring reliable cooperative perception, even under noisy pose conditions \cite{gu2023feaco}. Additionally, aerial monocular 3D object detection has been explored for improved perception from aerial perspectives, which can complement ground-based autonomous vehicles \cite{hu2023aerial}.

Although the aforementioned studies on cooperative perception have investigated the impact of lossy communication \cite{li2023learning} and link latency \cite{9718315}, most existing works \cite{xu2022cobevt, 9682601} were conducted under an unrealistic communication channel with time-invariant links. Moreover, it is noteworthy that how to determine fusion link establishment among nearby CAVs is still an open problem. For example, \textit{Who2com} utilizes a multi-stage handshake communication mechanism to decide which agents' information should be shared \cite{liu2020who2com}, while \textit{When2com} exploits a communication framework to decide when to communicate with other agents \cite{liu2020when2com}. However, these approaches are based on simplistic proximity-based design, such as fixed communication ranges or predefined neighborhood structures. They cannot capture the dynamic nature of the wireless channel among nearby CAVs. In contrast, graph-guided networks for irregularly sampled time series \cite{zhang2022graph} and graph-based optimization techniques, such as maximum matchings in bipartite graphs \cite{hopcroft1973matching}, can model proximity as edges in a graph, allowing for more flexible and adaptive link establishment based on actual signal strength, bandwidth, and other resources, leading to improved network throughput.

In addition to link establishment for fusion, compressing the data shared in V2V networks is equally important to maximize network throughput. Taking into account the burden of transmitting point cloud data, CAVs exploit multiple cameras to collect the surrounding data. However, each camera can also produce a significant amount of image data. For example, Google's autonomous vehicle is capable of amassing up to 750 megabytes of sensor data per second \cite{hu2019dynamic}. Thus, we have to compress those data by reducing their spatial and temporal redundancy. To reduce spatial redundancy, the general approach is to convert the raw data from high-definition data into a 2D matrix representation and then apply image compression methods to compress the data. As for temporal redundancy of streaming data, video-based compression methods are utilized to predict the content of the enclosed frames, such as MPEG \cite{8945224}. Learning-based compression methods with data-driven tools exhibit superior performance over JPEG, JPEG2000 \cite{skodras2001jpeg}, and BPG \cite{kovalenko2023bpg}. Learning-based compression methods, like modulated autoencoders, capture higher-level representations of the data by encoding important grain features while discarding less significant details \cite{yang2020variable}. Furthermore, modulated autoencoders can flexibly allocate bits according to the available bandwidth without training multiple models for each bitrate, while traditional methods like JPEG and MPEG use fixed compression ratios that are not adaptable to time-varying dynamic channels.

Most existing works on communication frameworks for CAVs assume the wireless channel is time-invariant. However, given the uncertainty of channels, network throughput can be improved if we leverage flexible and adaptive link establishment. Fang \textit{et al.}'s method improves sensing quality and coverage but does not fully consider the effects of throughput maximization on sensing performance and lacks analysis of latency and compression costs in their fusion model, thus limiting its comprehensiveness \cite{fang2024pacp}. Departing from prior research, this paper introduces a channel-aware scheme to optimize throughput for CAV data fusion, employing a self-supervised autoencoder for effective data compression. Key contributions of this work include:

\begin{itemize}
  \item To the best of our knowledge, this is the first contribution to leverage channel-aware throughput maximization to enable CAV data fusion. We formulate the problem as a mixed integer programming (MIP), which can be decomposed into two sub-problems. Then, we obtain closed-form solutions to the sub-problems for optimal data rate and compression ratio, respectively.
  \item We leverage a novel self-supervised autoencoder to alleviate the bottleneck of communications by reducing temporal and spatial redundancies across multi-frame data. The autoencoder provides an adaptive solution to striking a balance between data reconstruction accuracy and channel-aware compression ratio in V2V networks. Besides, the performance of reconstruction can be further enhanced by exploiting historical information through the fine-tuning strategy, which saves \textbf{42.0\%} of spectrum resources.
  \item With the OpenCOOD platform \cite{xu2022opv2v}, we conduct extensive experiments to verify the effectiveness of our proposed algorithm by comparing it with the state-of-the-art CAV's fusion schemes. Under the same wireless channel conditions, both the throughput and the average precision of the Intersection over Union (AP@IoU) show significant improvements, by at least \textbf{20.19\%} and \textbf{9.38\%}, respectively.
\end{itemize}

The remainder of this paper is organized as follows. The system model is detailed in Section \ref{sec:System_model}. In Section \ref{sec:Problem Formulation and Analysis}, we formulate the problem as an MIP problem and decompose it into two sub-problems. The design of our novel self-supervised autoencoder for adaptive compression is investigated in Section \ref{sec:Adaptive Compression Scheme}. In Section \ref{PERFORMANCE EVALUATION}, we provide the performance analysis of the proposed schemes in terms of throughput and IoU, followed by some related work in Section \ref{sec:RELATED WORK}. Finally, Section \ref{sec:Conclusion} concludes the paper.

\section{System Model}
\label{sec:System_model}
\begin{figure}[t]
  \centering
  \includegraphics[width=0.50\textwidth]{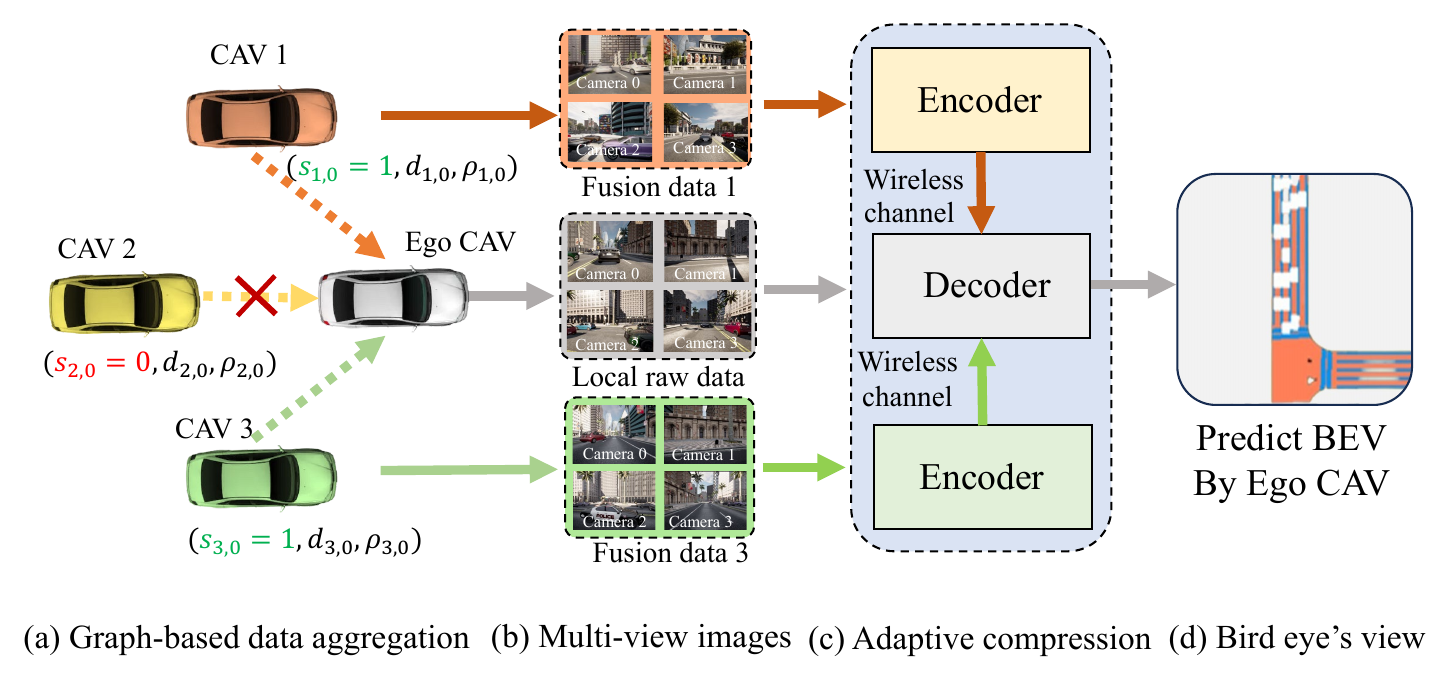}
  \caption{An example of a vehicle-to-vehicle (V2V) network consisting of four connected autonomous vehicles (CAVs). Fig. \ref{fig:system model}(a): CAV0 is the ego vehicle that can incorporate the viewpoints of CAV1 and CAV3. However, the link between CAV0 and CAV2 is disconnected to avoid negative impact on the overall throughput. Fig. \ref{fig:system model}(b): Each CAV collects the traffic status by four cameras. Fig. \ref{fig:system model}(c): Nearby CAVs encode camera data and then transmit them to the Ego CAV for decoding images. Fig. \ref{fig:system model}(d): The ego CAV predicts the bird eye's view (BEV) by fusing reconstruction data.}
  \label{fig:system model}
  \vspace{-3mm}
\end{figure}

Fig. \ref{fig:system model} illustrates a V2V network including four CAVs. Each CAV are equipped with sensing and communication modules, such as cameras and signal transmitter/ receiver. The role of cameras is to perceive the surrounding environment, and transmit the processed data to nearby CAVs through communication units. We assume that CAV0 is an ego vehicle, which makes decisions based on data collected from its four cameras as well as surrounding reachable CAVs. CAV1-CAV3 are nearby CAVs (or agents), sharing sensing data with CAV0 in terms of their surrounding environments. In so doing, CAV0 can observe invisible occlusions through other CAVs. Furthermore, CAV0 can run a throughput maximization scheme to determine the link establishment and transmission rate according to the acquired channel state information (CSI). As shown in Fig. \ref{fig:system model}(a), CAV1 and CAV3 are allowed to connect to the ego vehicle CAV0, while CAV2 is disconnected from CAV0. In Fig. \ref{fig:system model}(b), each CAV transmits its local sensing data to CAV0 after link establishment. In Fig. \ref{fig:system model}(c), It can be observed that we leverage a self-supervised autoencoder in non-ego vehicles to adaptively compress raw data according to CSI. In Fig. \ref{fig:system model}(d), the ego vehicle decodes and fuses the compressed data to obtain the prediction of the bird eye's view (BEV)\footnote{The bird's eye view (BEV) refers to a top-down perspective of a vehicle and its surroundings, synthesized from multiple sensors, which provides more comprehensive coverage of the vehicle's environment.}.  

Let $\mathbf{G}=\left( \mathbf{V},\mathbf{E} \right) $ represent the topology of the considered V2V network, where $\mathbf{V}=\left\{v_{1}, v_{2}, \ldots, v_{n}\right\}$ denotes the set of CAVs, and $\mathbf{E}$ is the set of the links between CAVs. Moreover, according to the 3GPP standards for 5G\cite{8891313}, the V2V communication exploits the cellular vehicle-to-everything (C-V2X) with Orthogonal Frequency Division Multiplexing (OFDM). In OFDM, the total bandwidth $W$ is divided into $K$ orthogonal sub-channels. The capacity of each sub-channel $C_{ij}$ yields:
\begin{equation}\label{eq:capacity}
  \begin{aligned}
    C_{i j}=\frac{W}{K} \log_{2}\left(1+\frac{P_{t} h_{i j}}{N_{0}\frac{W}{K}}\right), 
  \end{aligned}
\end{equation}
where $C_{ij}$ denotes the capacity of the sub-channel from the $i$th transmitter to the $j$th receiver, $P_t$ the transmit power, $h_{ij}$ the channel gain from the $i$th transmitter to the $j$th receiver, and $N_0$ the noise power spectral density.

The collaborative perception among nearby CAVs can be modeled as sensing data aggregation under the constraints of limited sub-channels. However, the potential number of communication links can significantly surpass the upper bound of the number of the sub-channels. For instance, if $N$ CAVs are implemented with full connectivity, the maximum number of directional links could reach $N\left(N-1\right)$. To maximize the throughput, we should prioritize those links that contribute most significantly to the overall graph flow, while links with poor channel quality should not be allowed to connect. If $v_{j}$ is an ego vehicle, $s_{ij}$ denotes the directional link establishment decision of link $(i,j)\in \mathbf{E}$. It is noted that $s_{ij}$ is the element of the binary matrix $\mathbf{S}^{n \times n}$, with its diagonal elements to be zeros. If $s_{ij}=1$, $v_{i}$ can transmit sensing data to the ego vehicle $v_{j}$, while $s_{ij}=0$ means the link $(i,j)$ is disconnected. According to the upper bound of the number of sub-channels $K$, the connectivity matrix elements $s_{ij}$ yields:
\begin{equation}\label{eq:subchannel}
  \begin{aligned}
    \sum_{i=1, i\ne j}^N{\sum_{j=1}^N{s_{ij}}}\le K.
  \end{aligned}
\end{equation}
Let $\mathbf{D}=\left\{d_{ij} \right\}_{N \times N}$ represent the non-negative matrix of transmission rates, where $\forall (i,j) \in \mathbf{E}$. Its element $d_{ij}$ denotes the amount of data transmitted from vehicle $v_i$ to vehicle $v_j$ and subsequently processed at $v_j$. It is noted that $d_{ij}$ yields:
{\small
\begin{equation}\label{eq:transmission rate}
  \begin{aligned}
    \rho _{ij}d_{ij}\le \min \left (C_{ij},A_i \right ), 
  \end{aligned}
\end{equation}
}where $\rho _{ij}\in(0,1]$ is the adaptive compression ratio, which is determined by the compression algorithm in Sec. \ref{sec:Adaptive Compression Scheme}. $A_j$ is the size of local sensing data of $v_i$ per second. This inequality denotes that $d_{ij}$ should be less than the achievable data rate or the sensing rate of the local data available at vehicle $v_i$. Besides, it is noted that insufficient compression ratio results in degradation of perception data accuracy, while excessively high compression ratio results in inefficiency of throughput maximization. Therefore, we have the constraint of the compression ratio as follows:
{\small
\begin{equation}\label{eq:the constraint of the compression ratio}
  \begin{aligned}
    \mathbf{1}^{\top}\rho _{j,\min}\preceq \mathcal{P} _j\preceq \mathbf{1}^{\top}\rho _{j,\max},
\end{aligned}
\end{equation}
}where we have $ \mathcal{P} _{{j}}=\left[ \mathrm{\rho}_{{j}},\mathrm{\rho}_{2{j}},...,\mathrm{\rho}_{{Nj}} \right] ^{\top}, \mathcal{P} =\left[ \mathcal{P} _1,\mathcal{P} _2,...,\mathcal{P} _{{N}} \right]$. Given the surrounding data obtained through collaborative perception, we assume that sensed data from closer vehicles is more important for safety, where they deserve a higher level of accuracy for potential observation of nearby occlusion (we admit that in environments with blockage, closer vehicle may not have better view than a further vehicle. We will investigate blockage cases in the future). Therefore, we assume that the adaptive compression ratio for the link $(i,j)$ yields:
{\small
\begin{equation}\label{eq:the constraint of the compression ratio 2}
  \begin{aligned}
    \rho _{ij}e^{L_{ij}}\geqslant \eta,
\end{aligned}
\end{equation}
}where $L_{ij}$ denotes the normalized distance between $v_i$ and $v_j$, and $\eta$ is a constant ranging from 0 to 1, and its specific value depends on the level of priority assigned by the ego vehicle to the nearby targets. Furthermore, the link establishment and data transmission rate must be optimized under the energy consumption constraints. Firstly, the transmission power dissipation of $v_i$ for link $(i,j)$ is given by:
{\small
\begin{equation}\label{eq:transmission power}
  \begin{aligned}
    E_{ij}^{t}=\tau _{j}^{t}P_t s_{ij}, 
  \end{aligned}
\end{equation}
}where $\tau_{j}^{t}$ is the transmission duration for the data aggregation of the ego vehicle $v_{j}$, and $P_{t}$ is the transmission power of the vehicles. Secondly, we denote $F_{j}$ as the computational capacity of vehicle node $v_{j}$. The processed data of $v_{j}$, which includes all its local data $A_{j}$ and the data received from its neighboring nodes (represented by $\sum d_{ij}$), must satisfy the following constraint:
{\small
\begin{equation}\label{eq:computing power constraint}
  \begin{aligned}
    A_j+\sum_{i=1,i\ne j}^N{\rho _{ij}s_{ij}d_{ij}}\leqslant F_j/\beta, 
  \end{aligned}
\end{equation}}
where $F_{j}$ and $\beta$ represent the local computational capacity and the ratio of required CPU cycles to the unit size of the input data, respectively. Thus, $F_j/\beta$ denotes the total size of processing data per second. Besides, $\beta$ is determined by the structure of the neural networks used in these scenarios, such as the self-supervised autoencoder. The computational energy consumption of $v_j$ can be obtained by:
{\small
\begin{equation}\label{eq:computing power}
  \begin{aligned}
    E_{j}^{c}=\epsilon _j\left( A_j+\sum_{i=1,i\ne j}^N{\rho _{ij}s_{ij}d_{ij}} \right)\tau_{j}^{c},
  \end{aligned}
\end{equation}
}where $\epsilon_{j}$ represents the energy consumed by CAV $v_{j}$ per unit of input data processed by its processing unit. $\tau_{j}^{c}$ denotes the data processing duration. By enforcing the constraint on total energy consumption, the CAV system can effectively balance the energy usage between computation and communication tasks, enabling CAVs to operate efficiently and prolonging their operational lifespan. Therefore, the total energy consumption of the CAV system needs to satisfy the constraint:
{\small
\begin{equation}\label{eq:power constraint}
  \begin{aligned}
    \sum_{i=1,i\ne j}^N \left(E_{ij}^{t}+E_{ij}^{c} \right) \leq E^{T}_j, \quad (j=1,2,\cdots,N)
  \end{aligned},
\end{equation}
}where $E^{T}_j$ denotes the sum of the energy dissipation threshold for the $j$th considered V2V network, including the ego vehicle $v_j$ and its nearby CAVs.

\section{Problem Formulation}
\label{sec:Problem Formulation and Analysis}

Based on the above system model, we next provide a formal description of the throughput maximization problem.  Throughput is one of the indicators for CAV scenario in terms of the preservation of perception accuracy and ensuring safety. To address these objectives, we focus on optimizing three key matrix variables: link establishment $\mathbf{S}$, data transmission rate $\mathbf{D}$, and compression ratio $\mathcal{P}$. High network throughput in CAVs ensures seamless communications between vehicles and the underlying network infrastructure, facilitating efficient data exchange for perception, decision-making, and coordinated actions. Let $T_{\mathrm{sum}}(\mathbf{S}, \mathbf{D})$ denote the whole data processing throughput of the considered system, as obtained by:
{\small
\begin{equation}\label{eq:throughput}
  \begin{aligned}
    T_{\mathrm{sum}}(\mathbf{S}, \mathbf{D})=\sum_{j=1}^{N}\left(A_{j}+\sum_{i=1, i \neq j}^{N} s_{ij} d_{ij}\right)
  \end{aligned}.
\end{equation}
}
By combining the constraints and objective function Eq. (\ref{eq:throughput}), we formulate the throughput maximization problem as:
{\small
\begin{equation}\label{OP:P0}
  \begin{aligned}
    \mathbf{P}: \quad \!&\max \limits_{\mathcal{P} ,\mathbf{S},\mathbf{D}} \sum_{j=1}^{N}\left(A_{j}+\sum_{i=1, i \neq j}^{N} s_{ij} d_{ij}\right)\\
    \textrm{s.t.} \quad
    &(\ref{eq:subchannel}), (\ref{eq:transmission rate}), (\ref{eq:the constraint of the compression ratio}), (\ref{eq:the constraint of the compression ratio 2}), (\ref{eq:computing power constraint}), (\ref{eq:power constraint}).
  \end{aligned}
\end{equation}
}
It is noted that problem $\mathbf{P}$ is an MIP problem since the optimization variables for link establishment $\mathbf{S}$ is discrete while data dissemination $\mathbf{D}$ and compression ratio $\mathcal{P}$ are continuous. The MIP problem is generally known to be NP-hard\cite{9430759}. Due to the difficulties of coupling variables in this MIP problem $\mathbf{P}$, it is computationally hard to find the optimal solution when the V2V network scale is large. While the control and decision of CAVs are latency-sensitive, we have to conceive a real-time optimization solver to address the issue of finding optimal solution to the MIP problem $\mathbf{P}$. Therefore, we decompose problem $\mathbf{P}$ into two sub-problems, i.e., the first part is to obtain the optimal data transmission rate and compression ratio (Sub-problem $\mathbf{P}_1$) while the second part is to get the optimal link establishment decision by solving $\mathbf{P}_2$. 

\textbf{(1)} The first sub-problem $\mathbf{P}_1$ in the $n$th round: Given the current link establishment $\mathbf{S}^{(n-1)}$, we optimize the variable matrices of the adaptive compression ratio $\mathcal{P}$ and data transmission rate $\mathbf{D}$. Then, $\mathbf{P}_1$ can be formulated as follows: 
{\small
\begin{equation}\label{OP:P1}
  \begin{aligned}
    \mathbf{P_1}: \quad \!&\max \limits_{\mathcal{P},\mathbf{D}}\ \  T_{\mathrm{sum}}\left(\mathbf{S}^{(n-1)}, \mathbf{D}\right)\\
    \textrm{s.t.} \quad
    &(\ref{eq:transmission rate}), (\ref{eq:the constraint of the compression ratio}), (\ref{eq:the constraint of the compression ratio 2}),\\
    &(\ref{OP:P1}\textrm{a}): A_j+\sum_{i=1,i\ne j}^N{\rho _{ij}s_{ij}^{(n-1)} d_{ij}}\leqslant F_j/\beta,\\
    &(\ref{OP:P1}\textrm{b}): \sum_{i=1,i\ne j}^N \left(E_{j}^{t}\big|_{s_{ij}^{(n-1)}}+ E_{ij}^{c}\big|_{s_{ij}^{(n-1)}} \right) \leq E^{T}_j,
  \end{aligned}
\end{equation}
}where $j=1,2,\cdots,N$. The sub-problem $\mathbf{P}_1$ is a type of nonlinear programming (NLP), because of the nonlinear constraints (\ref{eq:transmission rate}), (\ref{OP:P1}\textrm{a}) and (\ref{OP:P1}\textrm{b})\footnote{The product of the decision variables $\rho _{ij}$ and $d_{ij}$ are nonlinear.}. For non-convex problems, global optimization methods like branch and bound, genetic algorithms, or simulated annealing might be more appropriate, but these methods can be more computationally intensive. Thus, we attempt to fully linearize the original problem. Let $\mathbf{U}=\mathcal{P} \odot \mathbf{D}=\left[ u_{ij} \right] _{N\times N}$, where $\odot$ denotes the Hadamard product and $u_{ij}=\rho _{ij}d_{ij}$. By doing so, we linearize the product term in the constraints, then $\mathbf{P}_1$ can be reformulated as follows:
{\small
\begin{equation}\label{OP:P1-2}
  \begin{aligned}
    \mathbf{P_{1-1}}: \quad &\max_{\mathbf{U},\mathbf{D}}\ \sum_{j=1}^N{\left(A_j+\sum_{i=1,i\ne j}^N{s_{ij}^{(n-1)}d_{ij}}\right)}\\
    \text{s.t.}\quad
    &(\ref{OP:P1-2}\text{a}): u_{ij} \le \min(C_{ij},A_i), \\
    &(\ref{OP:P1-2}\text{b}): \max(\rho_{j,\min},{\eta}{e^{-L_{ij}}}) \le \frac{u_{ij}}{d_{ij}} \le \rho_{j,\max},\\ 
    &(\ref{OP:P1-2}\text{c}): \sum_{i=1,i\ne j}^N{s_{ij}^{(n-1)}u_{ij}} \leqslant \min(\gamma_j ^{(n-1)},\varphi_j),
  \end{aligned}
\end{equation}
}where $ \gamma_j ^{\left( n-1 \right)}=\frac{E^T}{\epsilon _j\tau _{j}^{c}}-\frac{\tau _{j}^{t}P_t\sum_{i=1,i\ne j}^N{s_{ij}^{(n-1)}}}{\epsilon _j\tau _{j}^{c}}-A_j, \varphi_j =\frac{F_j}{\beta}-A_j$. It is noted that the constraint (\ref{OP:P1-2}\textrm{c}) can be derived from (\ref{OP:P1}\textrm{a}) and (\ref{OP:P1}\textrm{b}). Though we introduce a bilinear equality constraint $u_{ij}$, (\ref{OP:P1-2}\textrm{b}) is still nonlinear. However, the objective function of $\mathbf{P_{1-1}}$ is to maximize $\sum_{j=1}^N{\sum_{i=1,i\ne j}^N{s_{ij}^{(n-1)}d_{ij}}}$, and we obtain the following inequality according to (\ref{OP:P1-2}\textrm{b}):
{\small
\begin{equation}\label{OP:P1-2-inequality}
  \begin{aligned}
    \frac{u_{ij}}{\rho _{j,\max}}\le d_{ij}\le \frac{u_{ij}}{\max \left( \rho _{j,\min},\frac{\eta}{e^{L_{ij}}} \right)},
  \end{aligned}
\end{equation}
}which provides an upper bound on the optimal value of the original problem $\mathbf{P_{1-1}}$. In order to maximize $d_{ij}$, we can maximize its upper bound. Therefore, we have a relaxation of the original problem as follows:
{\small
\begin{equation}\label{OP:P1-3}
  \begin{aligned}
    \mathbf{P_{1-2}}: \quad \!&\max \limits_{\mathbf{U}}\ \ \sum_{j=1}^N\sum_{i=1,i\ne j}^N  \frac{s_{ij}^{(n-1)}u_{ij}}{\max \left( \rho _{j,\min},\frac{\eta}{e^{L_{ij}}} \right)}\\
    \textrm{s.t.} \quad
    &(\ref{OP:P1-2}\textrm{a}) \ \mathrm{and}\ (\ref{OP:P1-2}\textrm{c}).
  \end{aligned}
\end{equation}
}
Problem $\mathbf{P_{1-2}}$ is a general linear programming problem, which can be solved by the simplex method or interior-point methods. We assume that the optimal result of $\mathbf{P_{1-2}}$ is $u_{ij}^{(n)}$. Thus, the current optimal solutions of data transmission rate and adaptive compression ratio are $d_{ij}^{(n)}=u_{ij}^{(n)} \left[{\max \left( \rho _{j,\min},\frac{\eta}{e^{L_{ij}}} \right)}\right]^{-1}$ and $\rho_{ij}^{(n)}=u_{ij}^{}/d_{ij}^{(n)}$,
respectively. Given that the values of $d_{ij}$ can be taken at the boundaries of the feasible region, the optimal solution of the problem $\mathbf{P_{1-2}}$ equates to the optimal solution of the original problem $\mathbf{P_{1-1}}$.

\textbf{(2)} The second sub-problem $\mathbf{P}_2$ in the $n$th round: Given the adaptive compression ratio $\mathcal{P}^{(n)}=\left[\rho_{ij}^{(n)}\right]_{N\times N}$ and data transmission rate $\mathbf{D}^{(n)}=\left[d_{ij}^{(n)}\right]_{N\times N}$, we optimize the variable matrix of the link establishment $\mathbf{S}$. Then, $\mathbf{P}_2$ can be formulated as follows: 
{\small
\begin{equation}\label{OP:P2}
  \begin{aligned}
    \mathbf{P_{2}}: \quad \!&\max \limits_{\mathbf{S}}\ \  \sum_{j=1}^N{\left( A_j+\sum_{i=1,i\ne j}^N{s_{ij}d_{ij}^{(n)}} \right)}\\
    \textrm{s.t.} \quad
    &(\ref{OP:P2}\textrm{c}): \sum_{i=1,i\ne j}^N{\mathrm{\chi}_{ij}^{\left( n \right)} s_{ij}}\le E_{j}^{T}-\tau _{j}^{c}\epsilon _jA_j ,\\
    &(\ref{OP:P2}\textrm{b}): \sum_{i=1,i\ne j}^N{s_{ij} u_{ij}^{(n)}}\leqslant \varphi_j \ \mathrm{and}\  (\ref{eq:subchannel}), 
  \end{aligned}
\end{equation}
}where $\mathrm{\chi}_{ij}^{\left( n \right)}=u_{ij}^{\left( n \right)}\epsilon _j\tau _{j}^{c}+\tau _{j}^{t}P_t$ can be obtained by the inequality constraint (\ref{eq:power constraint}). Since the variables \(s_{ij}\) are binary, $\mathbf{P_{2}}$ is a maximal flow problem, which can be solved by adding or removing links to obtain a higher throughput, i.e., the Ford-Fulkerson algorithm\cite{9345766}. Specifically, as for the $n$th round, the associated link establishment $\mathbf{S}^{(n)}\gets \mathbf{S}^{(n)}\backslash\{s_{ij}\}$ if link $(i,j)$ decreases the network throughput (Removing the link). Otherwise, we have $\mathbf{S}^{(n)}\gets \mathbf{S}^{(n)}\cup{\{s_{ij}\}}$ (Adding the link). As for $\mathbf{P_{2}}$, we need to repeat adding/removing links until an optimal solution that satisfies all constraints is found, or a preset number of iterations is reached.


\textbf{Complexity analysis}: Regarding the difficulty of solving the mentioned problems, the initial problem $\mathbf{P}_{1}$ can be changed into a linear programming problem $\mathbf{P}_{1-2}$. Suppose there are $N$ general CAVs and one ego CAV. In that case, the time complexity for $\mathbf{P}_{1-2}$ is $\mathcal{O}(N^{3.5} L)$, where $L$ is the size of the input data, and the method used for optimization is the interior point method. With the $K$ constraint in (\ref{eq:subchannel}), $\mathbf{P}_{2}$ maintains a low complexity of $\mathcal{O}(K)$. In practical applications, when addressing these problems sequentially in Algorithm \ref{alg:tmac-throughput}, the total complexity is determined by the higher of the two. Therefore, in scenarios where the number of CAVs is large, the overall complexity of the algorithm is governed by $\mathcal{O}(N^{3.5} L)$.

\section{Adaptive Compression Scheme}

\begin{figure*}[t]
  \centering
  \includegraphics[width=2\columnwidth]{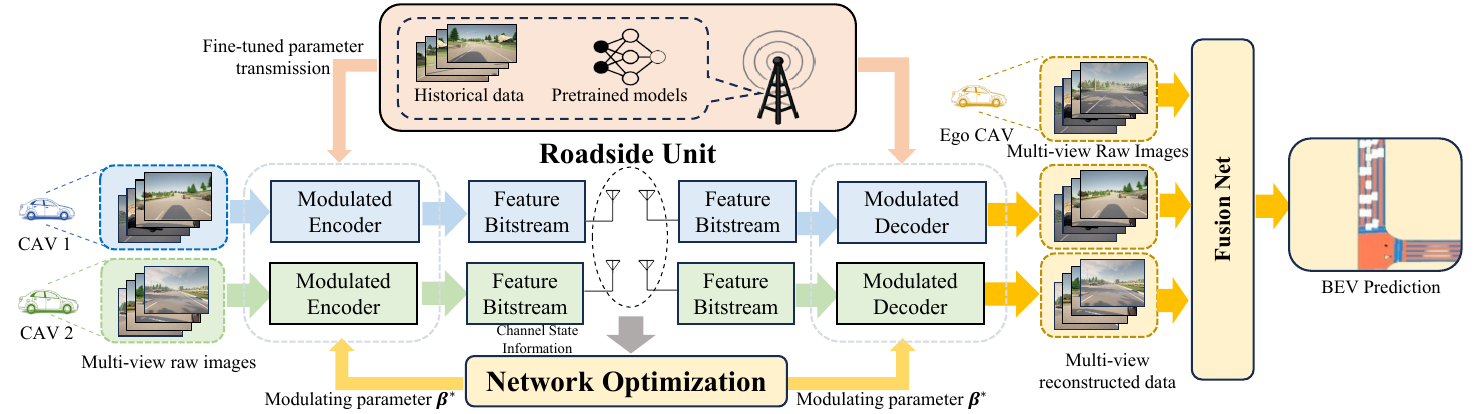}
  \caption{The overall architecture: 1) $\beta^{*}$ is obtained using Algorithm \ref{alg:tmac-throughput} based on the current channel conditions. 2) CAV1 and CAV2 fine-tune a small portion of historical images by updating parameters from roadside units. 3) CAVs use their encoders to convert images into a bitstream, which is then transmitted to the ego CAV. 4) The ego CAV can decode the received bitstream to reconstruct the images, while the reconstructed images are fused together in a Fusion Net to obtain BEV prediction.}
  \label{fig:TMAC}
  \vspace{-3mm}
\end{figure*}

\label{sec:Adaptive Compression Scheme}
The optimal solution obtained for compression in previous section is under fixed network topology and channel models. However, as the network evolves with time, topology and channels are subject to change,  and hence the compression scheme must be adapted. 
In this section, we introduce our proposed adaptive compression scheme, building upon the optimization results from Section \ref{sec:Problem Formulation and Analysis}. Firstly, we investigate traditional compression algorithms and the currently popular deep learning-based autoencoders in Sec. \ref{Background}. In Sec. \ref{Deep Learning-Based Compression Scheme}, we conceive an adaptive rate-distortion (R-D) trade-off scheme to dynamically adjust the obtained compression ratio $\mathcal{P}$. In Sec. \ref{Fine-tuning}, a fine-tuning based scheme was proposed to further eliminate the temporal redundancy of CAV data. Finally, we summarize our proposed adaptive compression scheme with the whole architecture of the compression network.
\subsection{Current Compression Schemes}
\label{Background}

Traditional lossy image compression techniques such as JPEG typically adopt a two-step paradigm, encompassing an encoding and decoding procedure, which are given by: 

(1) \textbf{The first stage is the encoding process}, where the input image, denoted as $i\in \mathbb{R}^{N}$, is transformed into a latent representation $h$ using a specific algorithm such as the Discrete Cosine Transform (DCT), which is represented mathematically as $h = f(i)$. To further minimize the data volume, a quantizer $Q$ is then applied to transform the continuous $h$ into a discrete vector $q\in \mathbb{H}^{D}$, following the relationship $q = Q(h)$. In the V2V scenario, $q$ is subsequently binarized and serialized into a bitstream $b$ for transmission. To optimize the transmission further, additional entropy coding techniques are employed to eliminate data redundancy, thus allowing for higher spectrum efficiency. 

(2) \textbf{The second stage is the decoding stage.} The received bitstream $b$ or the discrete vector $q$ undergoes a series of inverse transformations, including dequantization represented as $\hat{h} = Q^{-1}(q)$ and a reconstruction function defined as $\hat{i} = H(\hat{h})$. These processes aim to recover the original image from the compressed data, leading to the reconstructed image as the final output.

While traditional lossy image compression techniques have their merits, they often fall short when dealing with the dynamic and complex scenarios of V2V networks, where topology and channel conditions are subject to high variability. These traditional methods, heavily relying on static algorithmic solutions, struggle to adapt to these changing conditions, often resulting in compromised image quality and spectrum inefficiency. 

Conversely, the advent of Deep Learning-Based Compression (DBC) presents an innovative alternative, capitalizing on the adaptive capability of deep learning algorithms. Unlike their traditional counterparts, DBC methods leverage learned parameters from vast training data from roadside units, enabling them to accommodate the dynamic changes inherent in V2V scenarios. 

\subsection{Deep Learning-Based Compression Scheme}
\label{Deep Learning-Based Compression Scheme}

In the DBC framework, both encoder and decoder consist of convolutional layers. The encoder efficiently transduces the input image into a latent representation $h=f(i;\theta)$, where the transformation parameters $\theta$ are learned from the training data. Correspondingly, the decoder utilizes another set of parameters $\xi$, learned from the training phase, to reconstruct the image as $\hat{i} = H(h;\xi)$. The training process is guided by the minimization of the following expression:
\begin{equation}\label{eq:basic R-D equation}
  \begin{aligned}
    \mathop{\arg\min}_{\theta,\xi} R\left(b\right)+\beta D\left(i,\hat{i}\right),
  \end{aligned}
\end{equation}
where we follow the description in \cite{yang2020variable} that $R(b)=\mathbb{E} \left[ -\log _2\mathrm{Pr}\left( b \right) \right] $ denotes the bitrate, and $\mathrm{Pr}\left( b \right)$ can be estimated by entropy model. Besides, the distortion $D\left(i,\hat{i}\right)= \mathbb{E}\left [ \left \| i-\hat{i} \right \|^{2}  \right ]$ and a fixed parameter $\beta$ balances the trade-off between bitrate and distortion in the rate-distortion (R-D) tradeoff. Upon completion of the learning process involving gradient descent and backpropagation, the autoencoder acquires the learned parameters. For the sake of simplicity, we follow the assumption made in \cite{yang2020variable} and rewrite Eq. (\ref{eq:basic R-D equation}) as:
\begin{equation}\label{eq:trained R-D}
  \begin{aligned}
    \mathop{\arg\min}_{\theta,\xi}R\left(\hat{h};\theta \right)+\beta D\left(i,\hat{i}; \theta, \xi \right),
  \end{aligned}
\end{equation}
where the bitrate now defined as $R\left(\hat{h};\theta \right) = \mathbb{E}\left [ -\log_{2}\mathrm{Pr}\left( \hat{h}\right)  \right ]$, and the distortion represented as $D\left(i,\hat{i}; \theta, \xi \right) = \mathbb{E}\left [ \left \| i-\hat{i} \right \|^{2}  \right ]$.
It is noted that the fixed tradeoff parameter $\beta$ makes it hard to adapt to the dynamic change during V2V collaboration. Specifically, fixed $\beta$ can severely affect the effective decoding of compressed data at the receiver, resulting in perceptual degradation of the reconstructed image, thereby escalating the risk of collisions and other safety hazards. Therefore, it is necessary to adaptively adjust $\beta$ to ensure the reliability of inter-vehicle communications. In this subsection, we demonstrate the design of a DBC scheme with the dynamic adjustment of the tradeoff parameter $\beta$, effectively catering to the high volatility and complexity of V2V collaborative environments.
\addtolength{\topmargin}{0.06in}
More specifically, according to Sec. \ref{sec:Problem Formulation and Analysis}, we indicate that the optimal compression rate for the current CSI maintains a transformative relationship with the tradeoff parameter $\beta$ and the compression rate $\rho$:
\begin{equation}\label{eq:trained distortion}
  \begin{aligned}
    \beta_{i,j}= G\left( \rho_{i,j}  \right), \quad \text{s.t.} \quad\rho_{i,j}\propto \beta_{i,j},
  \end{aligned}
\end{equation}
where $G(\cdot)$ is a nonlinear function. For simplicity, $\rho$ denotes $\rho_{i,j}$ and $\beta$ denotes $\beta_{i,j}$. This implies that dynamically modifying the compression rate $\rho$ requires a corresponding adaptive adjustment of $\beta$.

Moreover, we reframe the traditional fixed rate-distortion problem with a control function $I\left(\beta\right)$ as a multi-rate-distortion problem to address the requirement of adaptability:
\begin{equation}\label{eq:multi R-D}
  \begin{aligned}
    \mathop{\arg\min}_{\theta,\xi}\sum_{\beta \in\Theta }^{ } R\left(\hat{h};\theta,\beta \right)+I(\beta) D\left(i,\hat{i}; \theta, \xi,\beta \right),   
  \end{aligned}
\end{equation}
where the set of tradeoff parameter is $\Theta=\left \{ \beta^{1},\cdots ,\beta^{M} \right \} $. The control function is introduced to amplify the importance of a specific rate-distortion (R-D) operating point \(\beta^{*}\) and allow for adaptive adjustment of the R-D tradeoff according to real-time CSI:
\begin{equation}\label{eq:importance control}
  \begin{aligned}
     I\left(\beta\right)  = \begin{cases}
    \beta^{m}, & \text{if } \beta=\beta^{m}. \\
    0, & \text{otherwise}. \\
\end{cases}
  \end{aligned}
\end{equation}
Consequently, problem (\ref{eq:multi R-D}) becomes:
\begin{equation}\label{eq:optimized multi R-D}
  \begin{aligned}
    \mathop{\arg\min}_{\theta,\xi} R\left(\hat{h};\theta,\beta^{*} \right)+\beta^{*} D\left(i,\hat{i}; \theta, \xi,\beta^{*} \right). 
  \end{aligned}
\end{equation}
This approach permits the adaptive adjustment of the R-D tradeoff according to real-time CSI by $I\left(\beta^{*}\right)$. The integration of the above adaptive compression (Sec. \ref{Deep Learning-Based Compression Scheme}) and network optimization (Sec. \ref{sec:Problem Formulation and Analysis}) can maximize throughput, called the \textbf{T}hroughput \textbf{M}aximization with \textbf{A}daptive \textbf{C}ompression (TMAC) algorithm as shown in Algorithm \ref{alg:tmac-throughput}. The TMAC algorithm enables the encoders in nearby CAVs and the decoder in the ego CAV to learn efficient image representation.
\begin{algorithm}[t]
    \caption{TMAC: Throughput Maximization with Adaptive Compression}
    \label{alg:tmac-throughput}
    \begin{algorithmic}[1]
      \REQUIRE Input the number of vehicles: $N$. The channel constraints: $K$, $C_{ij}$. The device parameters: $\rho_{j,\min}$, $\rho_{j,\max}$, $\eta$, $\tau_j^t$, $\tau_j^c$, $E_j^T$, $F_j /\beta$.
      \ENSURE Output the optimal compression ratio $\mathcal{P}$ , link establishment $\mathbf{S}$, data rate $\mathbf{D}$,  modulated parameter $\beta^{*}$, encoder, and decoder.
      \STATE Initialize the link establishment decision $s_{ij}^{(0)}$ as an $N \times N$ matrix of zeros;
      \STATE \% Find top $K$ largest capacity in each V2V network (excluding diagonal) and set $s_{ij} = 1 $;
      \FOR{$j$ from $0$ to $N-1$}
        \STATE Sort column in descending order and get the indices of the largest $K$ elements, store these indices to $i$;
        \STATE Set $s_{ij}$ elements at indices $(i,j)$ to $1$;
      \ENDFOR
      \WHILE {True}
        \STATE Solve the linear programming problem $\mathbf{P_{1-2}}$;
        \STATE Solve the maximal flow problem $\mathbf{P_{2}}$;
        \IF {the problem is infeasible}
            \STATE Find the connected link with minimum capacity, and remove it, i.e., $\mathbf{S}^{(n)}\gets \mathbf{S}^{(n)}\backslash\{s_{ij}\}$;
        \ELSE
            \STATE Search the other link to increase the whole throughput and add it, i.e., $\mathbf{S}^{(n)}\gets \mathbf{S}^{(n)}\cup{\{s_{ij}\}}$;
            \STATE Break the loop and report the solution;
        \ENDIF
      \ENDWHILE
      \STATE Obtain the optimal solution for $\mathcal{P} ,\mathbf{S},\mathbf{D}$.
      \STATE Calculate the tradeoff parameter $\beta^{*}$ by Eq. (\ref{eq:trained distortion}).
      \STATE Modulated encoders (decoders) learns to compress (reconstruct) images under current channel conditions.
    \end{algorithmic}
\end{algorithm}

\subsection{Fine-tuning Compression Strategy}
\label{Fine-tuning}
\begin{algorithm}[t]
    \caption{Fine-tuning Compression Strategy}
    \label{alg:fine-tuning}
    \begin{algorithmic}[1]
      \REQUIRE Input Image Sequence: \(I\), pretrained model: \(M_{pre}\), fine-tuning dataset: \(D_{ft}\) and learning rate: \(\alpha\).
      \ENSURE Output Image Sequence: \(\hat{I}\).
      \STATE \(\rho^{1} = 1.0\) and \(\rho^{*} = \text{Output of Algorithm } \ref{alg:tmac-throughput}\);
      \FOR{\(j \gets 0\) to \(m\)}
        \STATE Transmit the first \(m\) images without compression for fine-tuning: \(i^{j}_{1\rightarrow \text{ego}} = i^{j}\);
        \STATE Fine-tuning the model using the lossless data: \(M_{ft} = \text{Finetune}(M_{pre}, D_{ft}, \alpha)\) and \(\vartheta_{ft} = \vartheta_{pre} - \alpha \nabla  \text{MSE}(D_{ft}, M_{ft})\);
      \ENDFOR
      \FOR{\(j \gets m+1\) to \(n\)}
        \STATE Transmit the remaining images with compression ratio \(\rho^{*}\): \(\hat{i}^{j} = \text{Compress And Transmit}(i^{j}, \rho^{*})\).
      \ENDFOR
    \end{algorithmic}
\end{algorithm}
To leverage temporal redundancy between consecutive frames in vehicle-to-vehicle collaborative perception tasks, we have devised a method to perform fine-tuning of the compressor network by utilizing a nominal fraction of real-time data as the training set. The backbone\footnote{The backbone refers to a pre-trained network that is used as a starting point or feature extractor for a new task. Our designed fine-tuning aided compression network mainly relies on \cite{yang2020variable} and \cite{balle2018variational}.} of our approach is to incorporate a Modulated Autoencoder (MAE), complemented by our proposed fine-tuning strategy. Specifically, $i$ represents the input image, $\hat{i}$ represents the output image, $\rho^{1}$ denotes the compression ratio set to 1, and $\rho^{*}$ represents the optimized compression ratio obtained from $\mathbf{P_{1-2}}$. Firstly, CAV1 transmits a small number of uncompressed images to the roadside unit for fine-tuning. Afterward, CAV1 transmits the remaining images to the ego vehicle, and these images are compressed. Regarding the fine-tuning process, a modest fraction of real-time data $D_{ft}$ is selected as the finetuning dataset for a pre-trained model $M_{pre}$. The discrepancy between the actual labels in the dataset $D_{ft}$ and the predictions made by the fine-tuned model $M_{ft}$ is gauged using the Mean Square Error (MSE) loss function MSE$(D_{ft}, M_{ft})$. The gradient descent optimization procedure with backpropagation updates the parameters of the fine-tuned model as follows:
\begin{equation}\label{eq:update}
\begin{aligned}
    \vartheta_{ft} = \vartheta_{pre} - \alpha\nabla  \text{MSE}(D_{ft}, M_{ft}),
\end{aligned}
\end{equation}
where $\vartheta_{ft}$ and $\vartheta_{pre}$ represent the parameters of the fine-tuned and pre-trained models, respectively, while $\alpha$ stands for the learning rate.
Conceptually, this fine-tuning strategy enables the model to exploit historical information from similar scenes, thereby enhancing the fidelity of future image reconstructions. From an information-theoretic standpoint, we consider $D_{p}$ to represent the historical data, and $D_{f}$ to represent future data. After using fine-tuned compression strategy, the actual uncertainty of reconstructed image can be defined by the conditional entropy as follows:
\begin{equation}\label{eq:mutual information}
\begin{aligned}
    H\left( D_{f}|D_{p}\right)=H(D_{f})-I(D_{f};D_{p}),
\end{aligned}
\end{equation}
where the entropy $H\left( D_f\right)$ denotes the actual uncertainty of reconstructed image without the historical information:
\begin{equation}
\label{eq:entrophy}
\begin{aligned}
    H\left( D_f\right) = - \sum \mathrm{Pr}\left( D_f\right) \log { \left[ \mathrm{Pr}(D_{f}) \right]},
\end{aligned}
\end{equation}
while the mutual information $I\left( D_f;D_p \right)$ can be given by:
\begin{equation}\label{eq:mutual information}
\begin{aligned}
    I\left( D_f;D_p \right) =\sum{\sum{\mathrm{Pr}\left( d_f,d_p \right)}}\log \left[ \frac{\mathrm{Pr}\left( d_f,d_p \right)}{\mathrm{Pr}\left( d_f \right) \mathrm{Pr}\left( d_p \right)} \right],
\end{aligned}
\end{equation}
which represents the reduction in uncertainty about predicting future data after understanding historical data. Through learning from the historical data $D_{p}$, this model increases the mutual information $I(D_{f};D_{p})$. Consequently, the reduction in uncertainty about the future data and the decrease in conditional entropy lead to an elevated accuracy in reconstructing $D_{f}$, since $H\left( D_{f}|D_{p}\right)<H\left( D_f\right)$ when $I\left( D_f;D_p \right)>0$. For a comprehensive explanation of the specific fine-tuning strategies, please refer to \mbox{Algorithm \ref{alg:fine-tuning}}. In this pseudocode, \(j\) is a loop variable used to traverse the image sequence. \(m\) is a threshold that determines which images are transmitted using lossless compression and which are transmitted using lossy compression. 

\subsection{Latency}
\label{latency analysis}

The latency of the proposed algorithm in transmitting a data packet is primarily composed of two parts: (1) arising from the transmission of uncompressed data that participates in fine-tuning, and (2) stemming from the transmission of compressed data. For the first part, the latency can be expressed as follows:
\begin{equation}\label{eq:latency1}
\begin{aligned}
    L = L_{up}+L_{down}+L_{ft},
\end{aligned}
\end{equation}
where $L_{up}$ signifies latency of uplink, $L_{down}$ represents latency of downlink, and $L_{ft}$ refers to latency of fine-tuning. For the second part, the latency can be expressed as follows:
\begin{equation}\label{eq:latency2}
\begin{aligned}
    \hat{L}  = \hat{L}_{up}+\hat{L}_{down}+\hat{L}_{inf},
\end{aligned}
\end{equation}
where $\hat{L}_{inf}$ indicates latency of inference. If a data packet consists of n frames, out of which i frames are utilized for fine-tuning, then the total delay can be formulated as follows:
\begin{equation}\label{eq:latency total}
\begin{aligned}
    L_{total} = i*L + (n-i)*\hat{L}. 
\end{aligned}
\end{equation}
To get the experiment results about the latency of our proposed algorithm, please refer to Fig. \ref{fig:latency} in Sec. \ref{PERFORMANCE EVALUATION} for more details.

Overall, the unique aspect of our proposed architecture is to combine a modulated autoencoder with a network optimization algorithm that dynamically tweaks the compression rate to achieve an optimal throughput level in response to varying communication conditions, thereby enabling better sensing accuracy of CAVs. It is noted that the backbone of the fusion network is based on CoBEVT\cite{xu2022cobevt}. Please refer to Fig. \ref{fig:TMAC} for more details of our proposed architecture.



\section{PERFORMANCE EVALUATION}
\label{PERFORMANCE EVALUATION}
In this section, we first have conducted extensive experiments to evaluate the network throughput in terms of different communication settings, such as bandwidth, transmission power, the number of accessed CAVs, and vehicle distribution. Then, we compare the performance of raw data reconstruction with or without fine-tuned compression strategy. Finally, we provide the predicted results of BEV and the associated IoU, which illustrate the performance of cooperative perception.
\subsection{Dataset and Baselines}
\label{Dataset and Baselines}
\textbf{Dataset:} We validate our method using a CAV simulated platform OpenCOOD with OPV2V dataset\cite{xu2022opv2v}. Notably, the OPV2V dataset is now the sole available source of camera-based image data for V2V collaborative perception. This dataset has 73 diverse scenes and numerous connected vehicles, 11,464 frames, and over 232,000 annotated 3D vehicle bounding boxes, collected from the CARLA simulator\cite{dosovitskiy2017carla}.

In the context of V2V collaborative perception, we select two state-of-the-art baselines and conduct a comparison using No Fusion strategy:

\textbf{Baseline 1:} This scheme is mainly based on the Distributed Multicast Data Dissemination Algorithm (\textbf{DMDDA}) proposed in \cite{lyu2022distributed}, which optimizes throughput in a distributed manner. For fair comparison, the transmission model and simulation settings are set to the same as ours, shown in Sec. \ref{sec:Problem Formulation and Analysis} and Sec. \ref{Simulation setup}.

\textbf{Baseline 2:} This scheme, namely Fairness Transmission Scheme (\textbf{FTS}), is mainly based on reference \cite{9681261}, which makes a fair sub-channel allocation according to Jain’s Network Starvation Fairness Index.

\textbf{Baseline 3:} This scheme, namely \textbf{No Fusion} scheme, denotes we only use one ego vehicle to collect the surrounding information without fusing nearby CAVs' camera data.

\subsection{Simulation setup}
\label{Simulation setup}
Our simulation parameters are in accordance with the 3GPP standard\cite{8891313}. Specifically, vehicles' communication range is set at 200 m and the default number of cooperative vehicles is $10$ with the upper bound of sub-channel number $K=4$. Unless otherwise stated, vehicle speeds range from $0$ to $50$ km/h, generated by CARLA simulator\cite{dosovitskiy2017carla}. The other default simulation settings: The transmit power is 8 mW for each CAV. The entire bandwidth W is 200 MHz over the V2V network. The local data $A_j$ is fixed as $40$ Mbits, with task computation complexity $\beta$ at 100 Cycles/bit. The CPU capacity $F_j$ ranges uniformly from $1$ to $3$ GHz\cite{lyu2022distributed}. The computing and transmission power threshold $E_j^T=1$ kW. Besides, all vehicles are uniformly distributed according to a six-lane highway spanning 200 meters. There are three lanes for vehicles traveling in the same direction. Without loss of generality, we disregard the redundant frames in our simulation, such as cyclic redundancy checks and Reed-Solomon code, etc.

\subsection{Experimental Results}
\label{Experimental results}

\begin{figure*}[t]
  \centering
  \subfigure[Throughput vs. CAV num.]{
  \includegraphics[width=4.2cm]{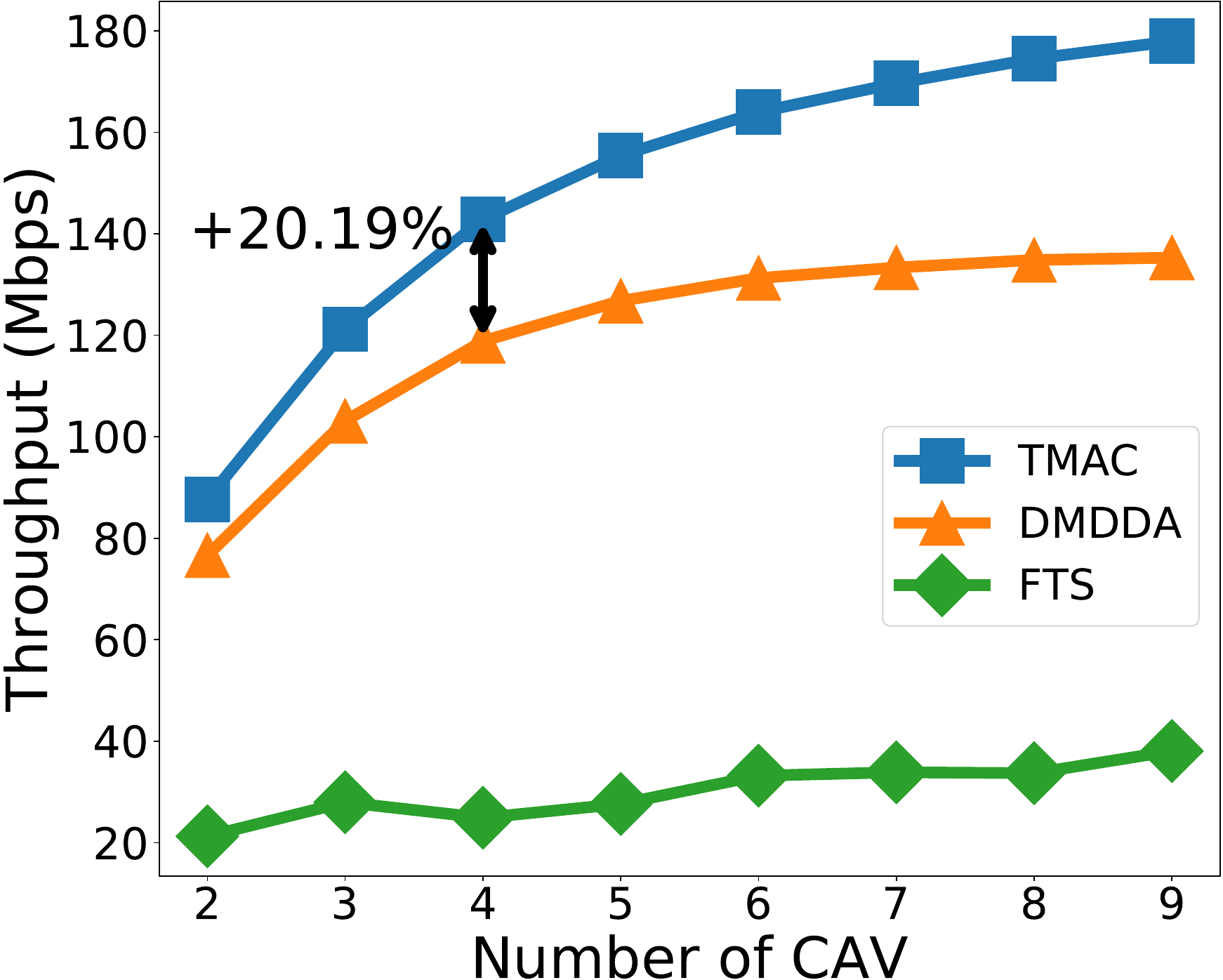}\label{fig:EE-TC-1}
  }
  \subfigure[Throughput vs. Bandwidth]{
  \includegraphics[width=4.2cm]{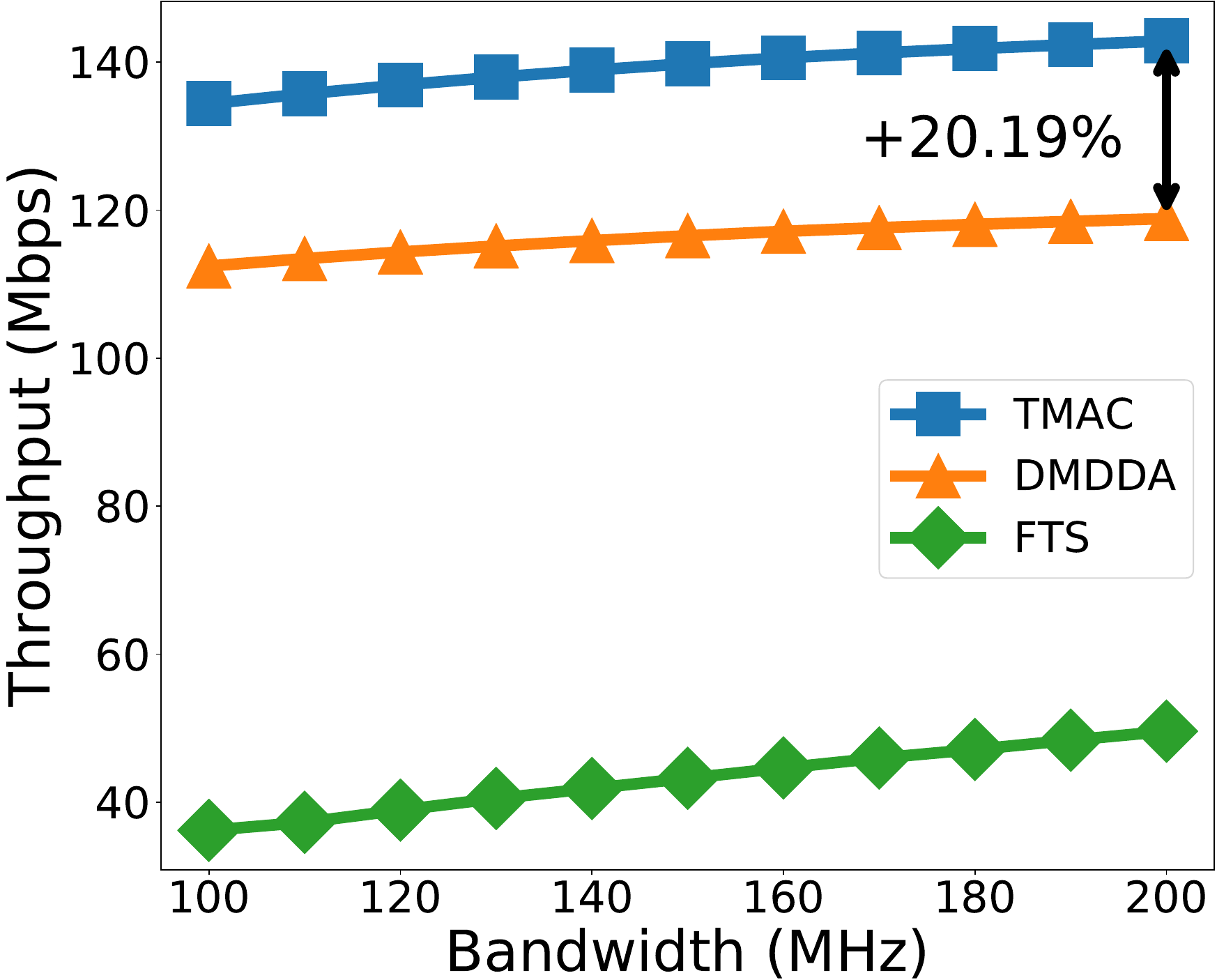}\label{fig:EE-TC-2}
  }
  \subfigure[Throughput vs. Power]{
  \includegraphics[width=4.2cm]{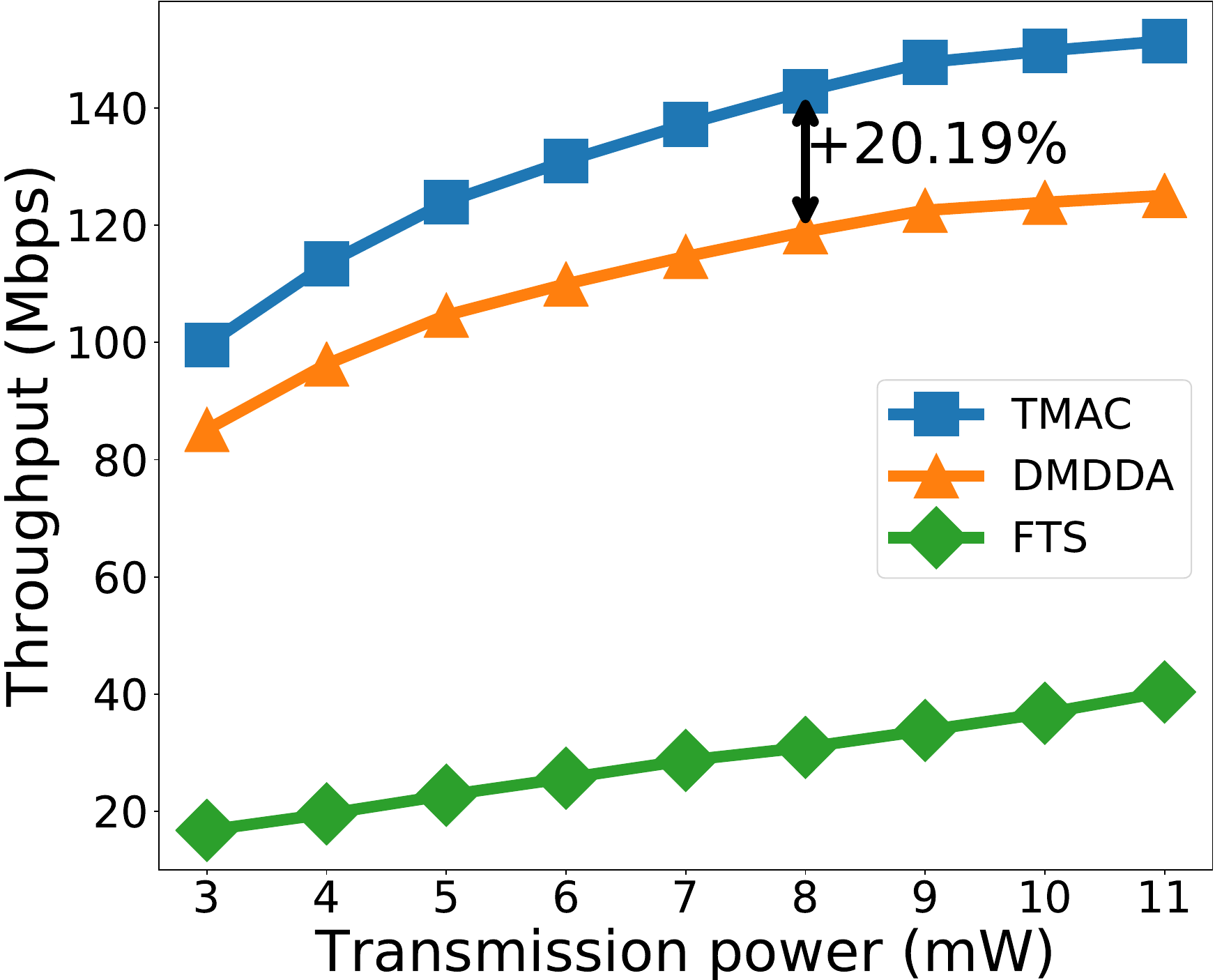}\label{fig:EE-TC-3}
  }
  \subfigure[Throughput vs. Max. Range]{
  \includegraphics[width=4.2cm]{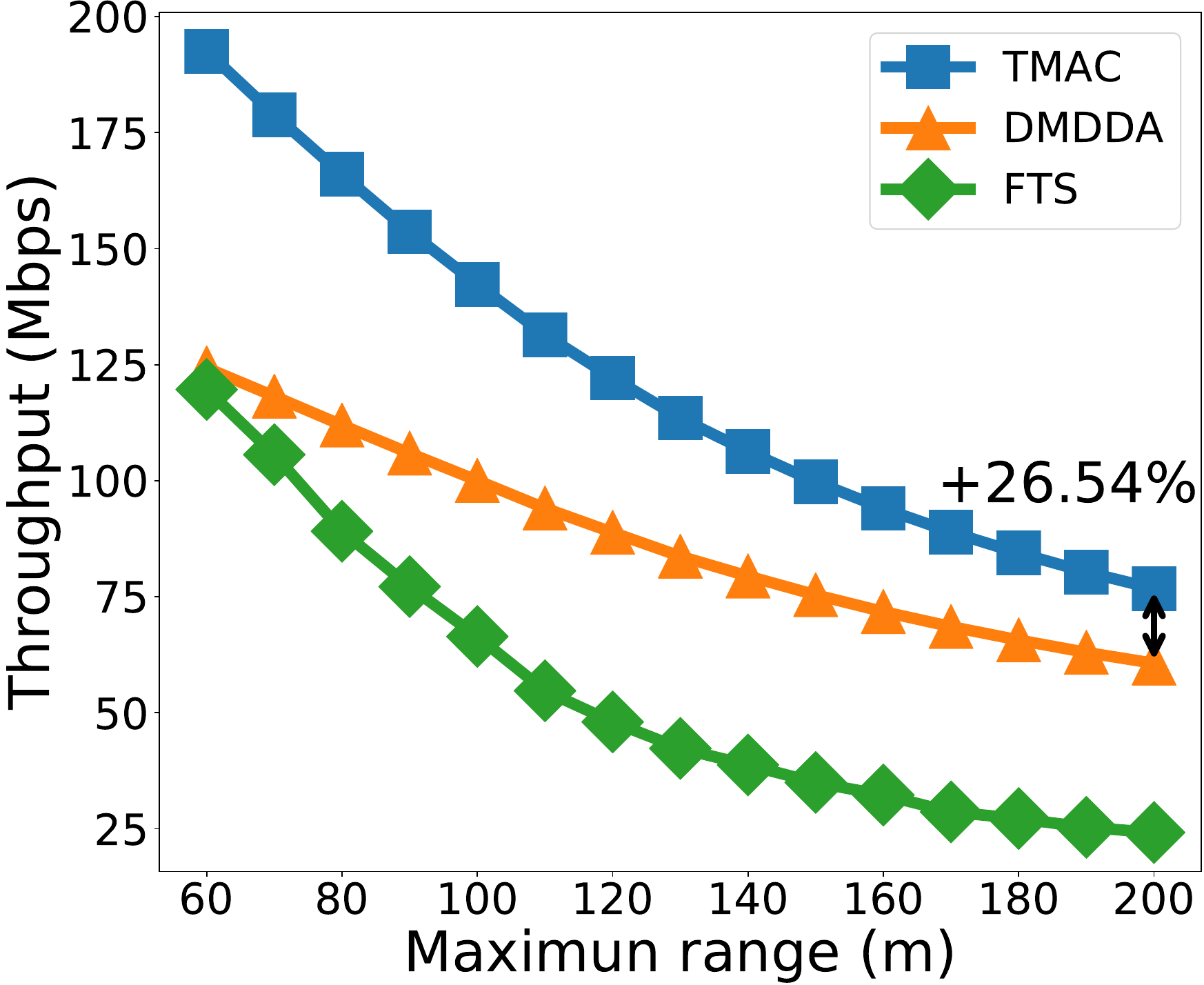}\label{fig:EE-TC-4}
  }
  \caption{Results in average network throughput under different communication parameters.}
  \label{fig:EE-MA}
  \vspace{-3mm}
\end{figure*}
We evaluate our proposed TMAC algorithm by comparing it with DMDDA and FTS. These two baselines represent common methods for V2V data transmission without dynamic adjustment of compression rates. Comparisons are performed under varying communication parameters, with each comparison test being conducted under identical conditions, as depicted in Fig. \ref{fig:EE-MA}. 
Firstly, we demonstrate the relationship between the number of cooperative vehicles and throughput in Fig. \ref{fig:EE-TC-1}. We observe that as the number of cooperating vehicles increases (ranging from 2 to 9), throughput also grows. However, as the upper limit of cooperating vehicles is close to the total number of vehicles, the growth trend in throughput slows down. The FTS strategy experiences performance degradation due to poorer communication channels of distant vehicles. Without adaptive compression, DMDDA cannot adjust data transmission size based on channel conditions. For cooperative perception involving four CAVs, the average throughput of V2V network increases by 20.19\% by relying on TMAC algorithm in comparison to DMDDA.

Fig. \ref{fig:EE-TC-2} highlights the direct correlation between the total bandwidth of the V2V network and throughput, with a bandwidth range of 100-200MHz. The throughput and bandwidth show an approximately linear relationship. Specifically, when $W=200$ MHz, the average throughput of V2V network increases by at least 20.19\% by relying on TMAC algorithm, compared with other baselines. In Fig. \ref{fig:EE-TC-3}, we present the impact of the transmission power of vehicle communication units on throughput. Fig. \ref{fig:EE-TC-4} illustrates the inverse relationship between the maximum range of vehicle distribution (ranging from 60 to 200 m) and throughput, i.e., as the range increases, throughput decreases. When the maximal range equals to 200 m, the average throughput of V2V network increases by at least 26.54\% by relying on TMAC algorithm in comparison to its competitors.

\begin{figure}[t]
  \centering
  \includegraphics[width=0.45\textwidth]{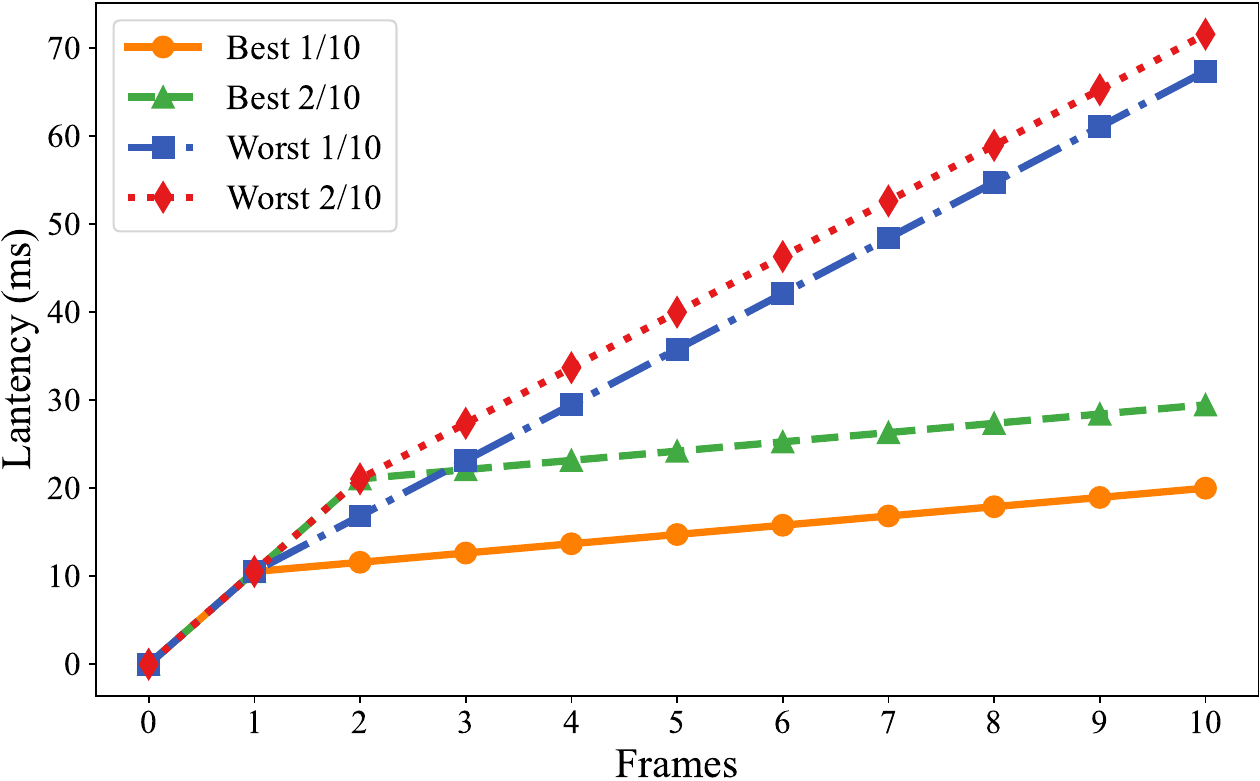}
  \caption{Latency of TMAC when transmitting a data packet in four different settings. $i/10$ denotes that the first $i$ ($i=1$ and $2$) frames out of ten is utilized for fine-tuning, with the remaining frames being compressed before being transmitted to the ego CAV. The term `best' represents the most favorable outcome achieved in our experiments, while `worst' refers to the least desirable result.}
  \label{fig:latency}
  \vspace{-6mm}
\end{figure}

In Fig. \ref{fig:latency}, we demonstrate the latency of TMAC when transmitting a data packet in four different settings. The latency test focuses solely on the transmission of image data, disregarding the transmission of control information due to its negligible volume. Given that the OPV2V dataset \cite{xu2022opv2v} captures ten frames per second, we configure each data packet to consist of ten frames. In this configuration, $i/10$ denotes that the first $i$ ($i=1$ and $2$) frames out of ten is utilized for fine-tuning, with the remaining frames being compressed before being transmitted to the ego CAV. The term `best' represents the most favorable outcome achieved in our experiments, while `worst' refers to the least desirable result. The latency composition of TMAC is mentioned in Sec. \ref{latency analysis}. For uplink and downlink, we follow the practical 5G-based V2X standard. For inference latency, we assume the use of a Tesla V100 for computations, and additional details on inference latency can be found in \cite{yang2021slimmable}. Our algorithm achieves a minimum latency of $19.99$ ms and a maximum latency of $71.53$ ms, both satisfying the requirement for transmission latency to be under $100$ ms \cite{zhang2018vehicular}.


\begin{figure}[t]
  \centering 
  \subfigure[Bandwidth saving vs. MS-SSIM]{
  \begin{minipage}[t]{0.5\linewidth}
    \centering 
  \includegraphics[width=1.65in]{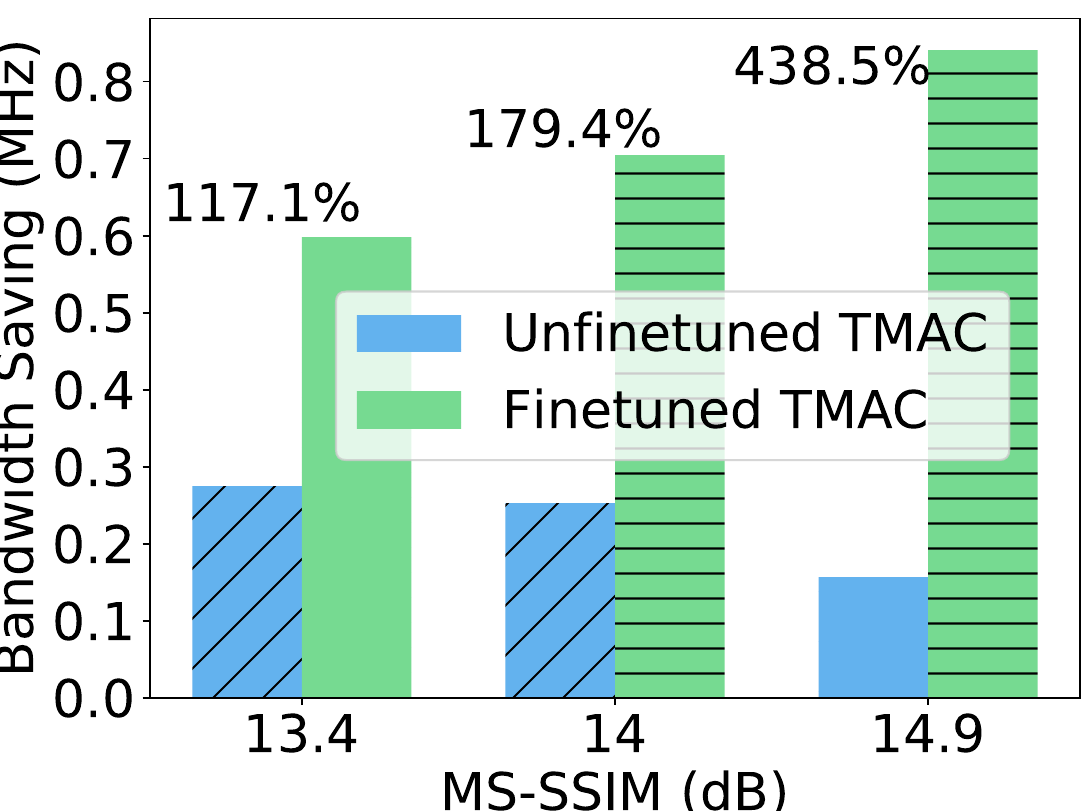}
  \end{minipage}%
  }%
  \subfigure[Bandwidth saving vs. PSNR]{
  \begin{minipage}[t]{0.50\linewidth}
    \centering 
  \includegraphics[width=1.65in]{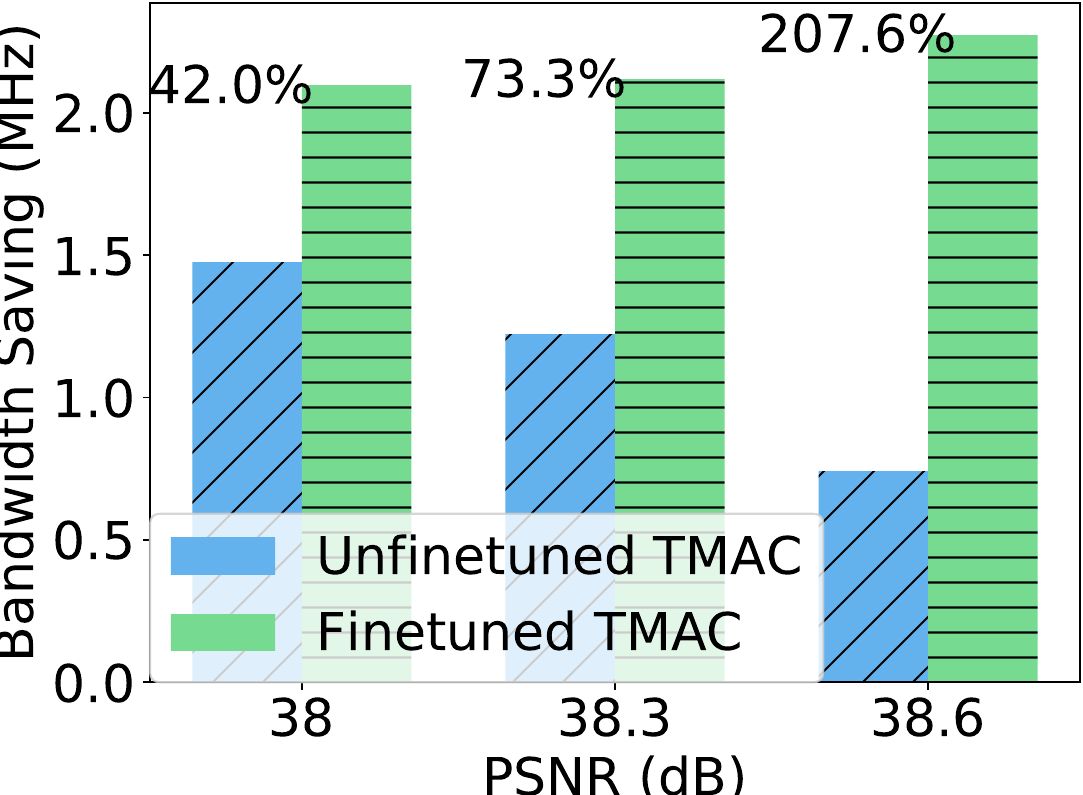}
  \end{minipage}%
  }%
  \flushleft 
  \caption{Comparison of bandwidth-saving performance between finetuned TMAC and unfinetuned TMAC at (a) the same MS-SSIM level, (b) the same PSNR level.}
  \label{fig:bandwidth saving}
  \vspace{-6mm}
\end{figure}

In Fig. \ref{fig:bandwidth saving}, despite preserving the same quality of image reconstruction (measured by both MS-SSIM (Multi-Scale Structural Similarity Index), which evaluates structural similarity across multiple scales, and PSNR (Peak Signal-to-Noise Ratio), which quantifies the signal-to-noise ratio affecting image quality), the fine-tuned TMAC requires less spectrum resources, which indicates a more efficient utilization of the spectral resources. Specifically, compared to the Unfinetuned TMAC, the Finetuned TMAC saves at least 42.0\% bandwidth. The superior performance of the fine-tuned TMAC can be attributed to the fact that finetuning allows the model to learn from historical data and leverage this information to reconstruct images more effectively.

\begin{table*}[]
\centering
\caption{The comparison of TMAC with three baseline methods under different parameters in terms of perception accuracy.}
\label{tab:table1}
\resizebox{1\textwidth}{!}{
\begin{tabular}{|c|ccc|ccc|ccc|}
\hline
\multirow{2}{*}{\diagbox[width=10em]{AP@IoU}{Parameters}} & \multicolumn{3}{c|}{\textbf{Power (mW)}} & \multicolumn{3}{c|}{\textbf{Bandwidth (MHz)}} & \multicolumn{3}{c|}{\textbf{Num. of CAVs}} \\ \cline{2-10}
 & \textbf{4} & \textbf{8} & \textbf{12} & \textbf{100} & \textbf{150} & \textbf{200} & \textbf{2} & \textbf{3} & \textbf{4} \\ \hline
\textbf{No Fusion} & 0.408 & 0.408 & 0.408 & 0.408 & 0.408 & 0.408 & 0.408 & 0.408 & 0.408 \\ \hline
\textbf{FTS} & 0.602 & 0.599 & 0.601 & 0.600 & 0.600 & 0.600 & 0.465 & 0.550 & 0.597 \\ \hline
\textbf{DMDDA} & 0.502 & 0.545 & 0.552 & 0.545 & 0.545 & 0.545 & 0.558 & 0.545 & 0.545 \\ \hline
\textbf{TMAC} & \textbf{0.607} & \textbf{0.653} & \textbf{0.651} & \textbf{0.655} & \textbf{0.653} & \textbf{0.656} &\textbf{ 0.661} &\textbf{ 0.656} & \textbf{0.653} \\ \hline
\end{tabular}
}
\end{table*}

Table \ref{tab:table1} presents the performance comparison of our proposed TMAC with three baseline methods under different power levels, bandwidths, and numbers of CAVs. The experimental results clearly demonstrate that our TMAC significantly outperforms FTS, DMDDA and No Fusion scheme in CAV recognition during the collaborative perception process. To ensure fairness in comparisons, we have set the default values of the parameters to be a power of 8 mW, a bandwidth of 200 MHz, and a total of 3 CAVs. Specifically, TMAC achieves superior AP@IoU scores, outperforming FTS by at least 9.38\% and DMDDA by 18.46\% in terms of the number of CAVs. The exceptional performance of our TMAC comes from the dynamic channel allocation by considering the importance of CAVs' data, thus greatly overcoming spectrum scarcity in real-world scenarios. In contrast, FTS's average channel resource allocation behavior results in resource deficiency for CAVs that require more resources, while allocating excess resources to CAVs that do not need as much. For DMDDA, the absence of compression implies that data transmission to the ego vehicle cannot be timely under limited spectrum bandwidth. Consequently, the Fuse Net may fuse data frames from different time instants, leading to performance degradation.

\begin{figure*}[t]
  \centering
  \includegraphics[width=2.03\columnwidth]{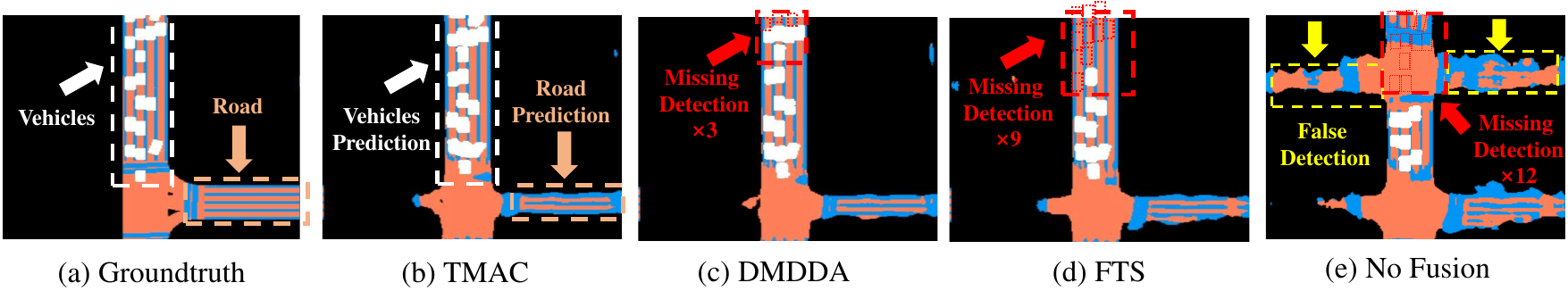}
  \caption{The BEV prediction of (a) Groundtruth, (b) TMAC, (c) DMDDA, (d) FTS, (e)No Fusion. The white blocks represent vehicles, and the remaining colored areas indicate the road. TMAC does not have false or missing detection, while DMDDA, FTS and No Fusion schemes do have several false and missing detection.}
  \label{fig:bev}
  \vspace{-3mm}
\end{figure*}

Fig. \ref{fig:bev} presents the BEV prediction for object detection across (a) Groundtruth, (b) TMAC, (c) DMDDA,  (d) FTS, and (e) No fusion. The Groundtruth (Fig. \ref{fig:bev}(a)) represents an ideal collaborative sensing communication. Notably, TMAC does not present any false or missing detections. In contrast, the perception results based on DMDDA and FTS algorithms, displayed in Figs. \ref{fig:bev}(c) and (d) respectively, exhibit several missing detections. The results from "No fusion" scheme with only one single vehicle perception, illustrated in \mbox{Fig. \ref{fig:bev}(e),} have numerous false and missing detections. The above results underscore the necessity of multi-view data fusion and the superiority of our proposed TMAC scheme.

\section{RELATED WORK}
\label{sec:RELATED WORK}
\textbf{V2V collaborative perception}: Vehicle-to-Vehicle Collaborative Perception combines the detector data from different CAVs through fusion networks, thereby expanding the perception range of each CAV and mitigating issues like blind spots \cite{hu2023adaptive, hu2024collaborative,hu2023towards}. For instance, Chen \textit{et al.} \cite{chen2019cooper} proposed the early fusion scheme, which fuses raw data from different CAVs, while Wang \textit{et al.} \cite{wang2020v2vnet} employed intermediate fusion, fusing intermediate features from various CAVs, and Rawashdeh \textit{et al.} \cite{rawashdeh2018collaborative} utilized late fusion, combining detection outputs from different CAVs to accomplish collaborative perception tasks. Although these methods show promising results in ideal conditions, in real-world environments where the channel conditions are highly variable, directly applying the same fusion methods often results in sub-optimal outcomes. 


\textbf{Throughput optimization}: High throughput can ensure more efficient data transmission among vehicles, thereby potentially improving the IoU of the cooperative perception system. Lyu \textit{et al.}\cite{lyu2022distributed} proposed a fully distributed graph-based throughput optimization framework by leveraging submodular optimization. Nguyen \textit{et al.}\cite{9204672} designed a cooperative technique aims to enhance data transmission reliability and improve throughput by successively selecting relay vehicles from the rear to follow the preceding vehicles. Ma \textit{et al.}\cite{8812911} developed an efficient scheme for the throughput optimization problem in the context of highly dynamic user request. However, the intricate relationship between throughput maximization and IoU has not been thoroughly investigated in the literature. This gap in the research motivates the need for more comprehensive studies that consider the role of throughput optimization in vehicular cooperative perception. 

\textbf{Camera data compression}: For V2V collaborative perception, data is usually gathered from either LiDAR or cameras. In this paper, we primarily use camera data for two key reasons. Firstly, camera data, within the same storage limits, provides higher resolution, crucial for accurate object recognition. Secondly, cameras are more cost-effective than LiDAR. Following data collection, participating vehicles compress their data before transmitting it to the ego vehicle to reduce transmission latency. However, existing collaborative frameworks often employ very simple compressors, such as the naive encoder consisting of only one convolutional layer used in V2VNet\cite{wang2020v2vnet}. Such compressors cannot meet the requirement of transmission latency under $100$ ms\cite{zhang2018vehicular} in practical collaborative tasks. Additionally, current views suggest that compressors composed of neural networks\cite{liu2018cnn} outperform compressors based on traditional algorithms\cite{david2012jpeg2000}. Nevertheless, these studies are typically focused on general data compression tasks and lack research on adaptive compressors suitable for practical scenarios in vehicle-to-vehicle collaborative perception. 


\section{Conclusions}
\label{sec:Conclusion}
In this paper, we have developed a channel-aware throughput maximization scheme for CAV cooperative perception. The proposed TMAC algorithm, combined with an adaptive compression scheme, enables us to dynamically adapt the compression rate for V2V transmissions under dynamic communication conditions, enhancing the performance of network throughput and perception accuracy. Additionally, we have also introduced a fine-tuning strategy to further eliminate spatial and temporal redundancies in the transmitted data. Experimentation on the OpenCOOD platform verifies the superiority of our algorithm compared to the existing state-of-the-art methods. The results demonstrate that TMAC can improve the network throughput by \textbf{20.19\%} and \textbf{2 times} over DMDDA\cite{lyu2022distributed} and FTS \cite{9681261}, respectively. Regarding perception accuracy (AP@IoU), TMAC outperforms DMDDA and FTS for BEV prediction with different number of CAVs by a minimum of \textbf{18.5\%} and \textbf{9.38\%}, respectively. Furthermore, after exploiting the historical information, the finetuned TMAC can save at least \textbf{42.0\%} of spectral resources and the optimal latency of our proposed algorithm achieved is \textbf{19.99 ms}.

\ifCLASSOPTIONcaptionsoff
  \newpage
\fi
\bibliographystyle{IEEEtran}

\bibliography{ref,ref2}

\end{document}